# Calibration Venus: An Interactive Camera Calibration Method Based on Search Algorithm and Pose Decomposition

**Wentai Lei [1], Mengdi Xu [1], Feifei Hou [1,\*], Wensi Jiang [1]**

[1] School of Computer Science and Engineering, Central South University
\* Correspondence: houfeifei@csu.edu.cn

**Abstract:** In many scenarios where cameras are applied, such as robot positioning and unmanned driving, camera calibration is one of the most important pre-work. The interactive calibration method based on the plane board is becoming popular in camera calibration field due to its repeatability and operation advantages. However, the existing methods select suggestions from a fixed dataset of pre-defined poses based on subjective experience, which leads to a certain degree of one-sidedness. Moreover, they does not give users clear instructions on how to place the board in the specified pose.

This paper proposes a new interactive calibration method, called Calibration Venus, that integrates two tasks of pose selection and user guidance. We develop a search algorithm that can select the optimal pose that minimizes estimation uncertainty in the entire pose space. In addition, we decomposing and displaying the pose, and then the user can follow the guide step by step to complete the placement of the calibration board. Experimental results evaluated on simulated and real datasets show that the proposed method can reduce the difficulty of calibration, improve the accuracy of calibration, and bring a good user experience.

## 1. Introduction

In recent years, due to the rise of virtual reality[1] and unmanned driving[2], higher requirements have been put forward for the perception of real scenes, which has promoted the rapid development of 3D reconstruction[3,4] and photogrammetry[5] technology. And camera calibration is a necessary part of many applications based on the camera, the quality of calibration greatly affects the effect of subsequent links. The purpose of camera calibration[6,7] is to obtain the internal camera geometric and optical characteristics (intrinsic parameters), including the focal length, optical center projection position and distortion coefficient. The general method[8,9] is to fix the camera, and obtain images of different positions and orientation by moving the calibration object. Then use the corresponding relationship between the feature points of the calibration object and its imaging points to realize the estimation of internal parameters. We use pose to represent the position and orientation of the calibration object relative to the camera, which can be specified by a rotation vector and a translation vector.

The calibration images obtained by the combination of different poses have very different effects when used for calibration. Now there have a few studies on the influence of the pose of the calibration object. Zhang[10] proposed that parallel poses that would cause degradation should be avoided. (the rotation vector of the two poses is exactly the same.), and the rotation angle should not be too large, because as the rotation angle increases, the feature point detection error will also increase. If the rotation angle is too small, the pose of the target will be almost parallel to each other;

Xie Zexiao et al.[11] studied the properties of the homography matrix describing the geometric transformation relationship between the calibration board and its imaging, and summarized that the angle parallel to the image plane should be avoided; Triggles[12] related the angle to the error in focal length. He also found that the rotation angle needs to be at least 5 degrees; Sturm and Maybank[13] separated the focal length and the optical center projection position for consideration, and they explored the possible singularities when using one and two calibration images of various methods. It is found that if each target orientation in the calibration set is parallel to the image plane, then the focal length cannot be determined; Rojtberg[14] associated the pose and the constraints of a single parameter together, and found that increasing sampling accuracy in image regions exhibiting strong distortions can effectively constrain the distortion coefficients, and maximizing the spread angle between image plane and calibration pattern can better constrain the focal length and the projection position of the optical center.

For users, how to choose the pose during the calibration process, that is, how to place the calibration object is an inevitable problem. Traditional calibration methods allow people to make their own decisions. It is difficult for users to complete this work without calibration experience, and even professionals have to go through many attempts to obtain a satisfactory calibration result, and a successful calibration is difficult to be reproduced. In response to these problems, interactive calibration has started to emerge. Unlike traditional calibration that let the user to choose the pose alone, interactive calibration will select each target pose for users, and help users arriving these poses in the actual calibration process.

Some recent research work of interactive calibration is as follows: Richardson[15] performed hypothetical calibration on 60 candidate poses, and selected the best result as the target pose. However, because the candidate poses are uniformly sampled within the camera's receptive field, the degradation and spread angle are not considered. They fill the calibration board with color during pose guidance, which can help users better position the calibrate board, but the user can only achieve the target pose by his own feeling. Rojtberg[14] associated pose and internal parameter constraints, so that the internal parameter with the largest variance in the calibration result can be specifically constrained in order to minimize the uncertainty of the calibration result and better constrain internal parameter. He also grouped the internal parameters and used different pose selection strategies. The focal length and the optical center projection position were set as a group, and the strategy of maximizing the spread angle was adopted; the distortion coefficients were a group, and the strategy was to increase the sampling accuracy of the region with the most distorted. When the pose is guided, they adopt the strategy of real-time detection of the calibration board, and visualize the coordinate axis of the calibration board under the current pose and the target pose, so that the user has a reference and the pose realization becomes a lot easier, but real-time detection also brings the problem of time delay. The above-mentioned methods all select poses from a set of preset candidate poses, which may cause some problems when applied to different camera models and calibration patterns. In addition, the aforementioned user guidance strategy requires a lot of effort for the user to consider how to spatially rotate the calibration board.

In response to these problems, this paper proposes a search algorithm based on the entire pose space to select the optimal pose set. Based on the work of Rojtberg[14], this paper designs a search algorithm to optimize the choice of pose. In specific implementation, after each round of calibration, the pose generated on his pose generation strategy is used as the initial solution, and the uncertainty of the internal parameter estimation is minimized as the loss function. An optimal pose is searched in pose set and is used as the expected pose for the next calibration. In addition, this article proposes a new pose decomposition guidance method: no longer based on the coordinate axis as a reference (real-time detection of the image is required), but a lazy loading processing[16] method to reduce the delay and improve the real-time performance, and decompose desired poses only when needed. The decomposition result can be understood as the posture after translation and rotation in the direction of each coordinate axis. Then the decomposed poses are projected and saved respectively, and displayed in order, which allows users to placing the calibration board in a step-by-step manner makes the entire guidance process more intuitive, efficient and faster.

The remaining chapters of this article are arranged as follows: Section 2 introduces the basic theory of camera calibration, Section 3 introduces the calibration process and various links of this method, Section 4 conducts an experimental evaluation of this method in simulation and actual measurement scenarios, Section 5 summarize and forecast the full text.

**2. Basic theory**

In many computer vision tasks that use camera, camera calibration is undoubtedly one of the most important pre-work[17]. The common method of camera calibration is to model the camera imaging process, and then estimate the camera internal parameters through data sampling. Interactive camera calibration is one of the current frontier researches in the field of camera calibration. The purpose is to design a calibration method that is user-friendly, easy to operate, stable and effective in the calibration process. The following article will introduce the principle of camera imaging and a classic traditional calibration method: Zhang's calibration method[10], the main difference between traditional calibration and interactive calibration method.

*2.1. Camera imaging principle*

The camera maps a 3D point in the real world to a pixel point on a two-dimensional image. The mapping process can be described by means of coordinate system transformation. The coordinate systems commonly used in camera imaging models are given below.

2.1.1. Coordinate system definition

**World coordinate system:** since the camera can be placed anywhere in the real environment, select a reference coordinate system in the environment to describe its position, which is called the world coordinate system

**Camera coordinate system:** a 3D rectangular coordinate system rectangular coordinate system established with the optical center of the camera as the origin and the optical axis as the positive half-axis of the Z axis. It is the coordinate system when the camera is standing on its own angle to measure objects

**Projection plane coordinate system:** a coordinate system established with the intersection of the camera's optical axis and the projection plane as the origin to indicate the physical position of the pixel

**Image coordinate system:** a coordinate system established based on camera imaging plane

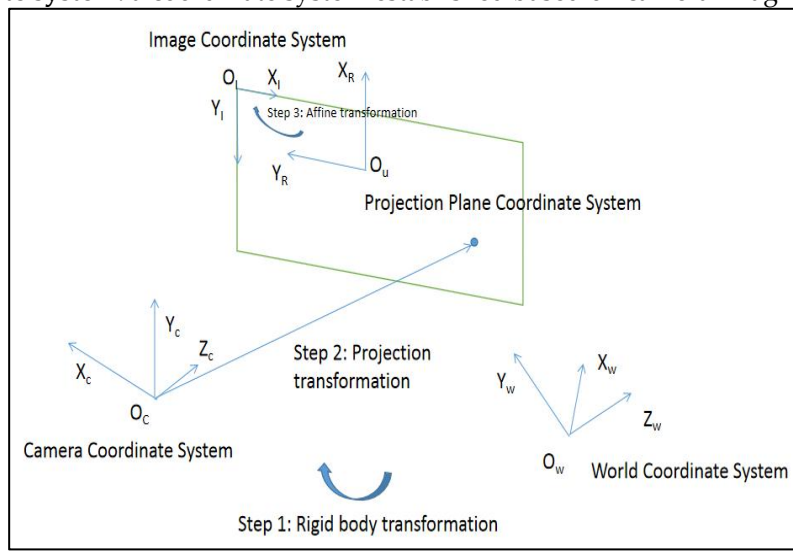

**Figure 1.** Coordinate system conversion diagram.

The camera imaging process can be described by the transformation relationship of the coordinate system. The transformation relationship of a simple camera model called the pinhole imaging model[18] is as follows.

2.1.2. Pinhole model

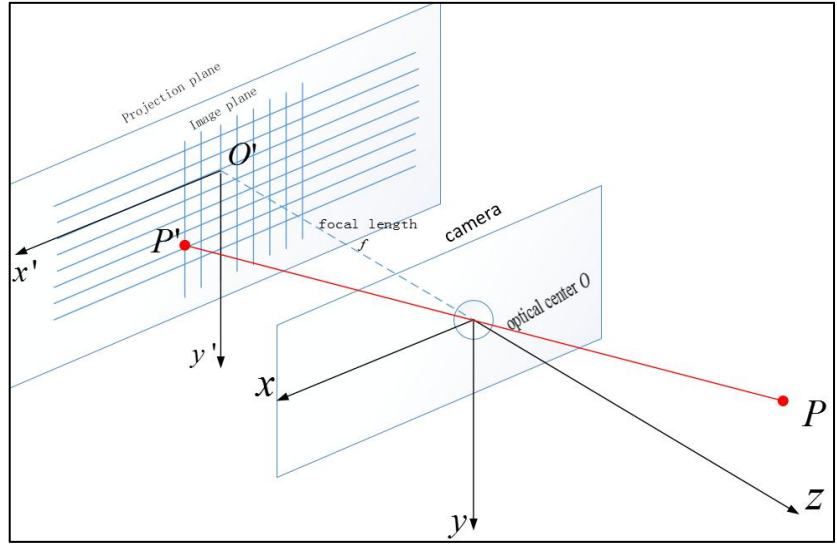

**Figure 2.** Schematic diagram of pinhole imaging model.

As shown in the Figure 2, in the camera coordinate system, the process of mapping a 3D point in the real world to a pixel on the image plane includes:
Transform from the world coordinate system to the camera coordinate system, where R and t are collectively referred to as external camera parameters

$$\begin{bmatrix} X_c \\ Y_c \\ Z_c \\ 1 \end{bmatrix} = \begin{bmatrix} R & t \\ 0 & 1 \end{bmatrix} \begin{bmatrix} X_w \\ Y_w \\ Z_w \\ 1 \end{bmatrix} \quad (1)$$

Transformed into the projection plane coordinate system after projection transformation

$$Z_c * \begin{bmatrix} x \\ y \\ 1 \end{bmatrix} = \begin{bmatrix} f & 0 & 0 \\ 0 & f & 0 \\ 0 & 0 & 1 \end{bmatrix} * \begin{bmatrix} X_c \\ Y_c \\ Z_c \end{bmatrix} \quad (2)$$

Transform to the image coordinate system through affine transformation

$$\begin{bmatrix} u \\ v \\ 1 \end{bmatrix} = \begin{bmatrix} \frac{1}{d_x} & 0 & u_0 \\ 0 & \frac{1}{d_y} & v_0 \\ 0 & 0 & 1 \end{bmatrix} * \begin{bmatrix} x \\ y \\ 1 \end{bmatrix} \quad (3)$$

Through the above conversion relationship, the conversion from a certain three-dimensional point in the world coordinate system to a pixel point in the two-dimensional image is realized. Due to the limitations of the inherent manufacturing process of the camera, the phenomenon that the pixels appear to deviate from their theoretical positions, this phenomenon is called distortion.

2.1.3. Distortion

The lens inside the camera will cause radial distortion[19] and tangential distortion[20] due to its shape and non-parallelism with the image plane during assembly. The process can be expressed by the following equation:

$$\Theta(x,y) = \begin{cases} x_{correct} = x(1 + k_1 r^2 + k_2 r^4 + k_3 r^6) + 2p_1 xy + p_2(r^2 + 2x^2) \\ y_{correct} = y(1 + k_1 r^2 + k_2 r^4 + k_3 r^6) + p_1(r^2 + 2y^2) + 2p_2 xy \end{cases} \quad (4)$$

After introducing distortion, the entire projection process can be simplified to:

$$\Lambda(K; R; t; Q) = K * \Theta(\frac{1}{Z_c}(R * Q + t)) \quad (5)$$

Where Q represents a three-dimensional point, K is the internal parameter matrix.

*2.2. Zhang's calibration method*

Camera calibration is the process of estimating the internal parameters of the camera, which will be used for subsequent computer vision tasks, so the quality of the camera calibration greatly affects the application effect of the camera-based scene.

The classic camera calibration method needs to capture images of a calibration object in multiple different poses (the position and angle of the calibration object relative to the camera), and use the homography relationship with its imaging to solve the internal parameters in a closed form, and then through a nonlinear optimization algorithm to correct the internal parameters. The following briefly introduces the classic Zhang's calibration method[10].

The process of mapping the feature points on the two-dimensional plane calibration board[21,22] to the pixels on the image is regarded as homography transformation (plane-to-plane transformation). That is, in the coordinate system with the calibration board as the plane and the direction perpendicular to the calibration board as the axis, the coordinate value of all the characteristic points is zero. Therefore, the projection equation can be simplified as:

$$\lambda \begin{bmatrix} u \\ v \\ 1 \end{bmatrix} = K \begin{bmatrix} r_1 & r_2 & t \end{bmatrix} * \begin{bmatrix} X \\ Y \\ 1 \end{bmatrix} \quad (6)$$

$r_1, r_2$ are the first two column vectors of the rotation matrix R, and $\lambda$ is the scaling factor.

*2.2.1. Closed-form solution*

Define symmetric matrix B in Equation 6. Let $b = \begin{bmatrix} B_{11} & B_{12} & B_{22} & B_{13} & B_{23} & B_{33} \end{bmatrix}^T$, where $B_{ij}$ is the element in row i and column j in B.

$$B = K^{-T} K^{-1} = \begin{bmatrix} \frac{1}{\alpha^2} & \frac{-\gamma}{\alpha^2 \beta} & \frac{v_0 \gamma - u_0 \beta}{\alpha^2 \beta} \\ \frac{-\gamma}{\alpha^2 \beta} & \frac{\gamma^2}{\alpha^2 \beta^2} + \frac{1}{\beta^2} & -\frac{\gamma(v_0 \gamma - u_0 \beta))}{\alpha^2 \beta^2} - \frac{v_0}{\beta^2} \\ \frac{v_0 \gamma - u_0 \beta}{\alpha^2 \beta} & -\frac{\gamma(v_0 \gamma - u_0 \beta))}{\alpha^2 \beta^2} - \frac{v_0}{\beta^2} & \frac{(v_0 \gamma - u_0 \beta)^2}{\alpha^2 \beta^2} + \frac{v_0^2}{\beta^2} + 1 \end{bmatrix} \quad (7)$$

Let Homography matrix $H = \begin{bmatrix} h1 & h2 & h3 \end{bmatrix} = K \begin{bmatrix} r_1 & r_2 & t \end{bmatrix}$. Since the rotation matrix R has the property that the column vectors are all unit vectors and are orthogonal to each other, the following two constraints can be obtained, which can be expanded and combined to obtain a homogeneous linear equation system:

$$\begin{cases} h_1^T K^{-T} K^{-1} h_2 = 0 \\ h_1^T K^{-T} K^{-1} h_1 = h_2^T K^{-T} K^{-1} h_2 \end{cases} \Rightarrow \begin{bmatrix} v_{12}^T \\ (v_{11} - v_{12})^T \end{bmatrix} b = 0 \quad (8)$$

with
$$v_{ij} = [h_{i1}h_{j1}, h_{i1}h_{j2} + h_{i2}h_{j1}, h_{i2}h_{j2},$$
$$h_{i3}h_{j1} + h_{i1}h_{j3}, h_{i3}h_{j2} + h_{i2}h_{j3}, h_{i3}h_{j3}]^T,$$
$h_{ij}$ is the element in row i and column j in H.

When n calibration images are observed, combining n such formulas can get VB=0, where V is a $2n \times 6$ matrix. When the number of images>2, the equation can calculate a solution of b.

2.2.2. L-M optimization

The obtained solution is used as the initial value, and the L-M method[23] is used for optimization, and then the internal parameter value can be calculated from the relationship between it and the internal parameter. The optimization goal is to minimize the following formula:

$$\sum_{i=1}^{n}\sum_{j=1}^{m}\|m_{ij} - \hat{m}(A, R_i, T_i, M_j))\|^2 \qquad (9)$$

*2.3. Traditional calibration and interactive calibration*

The traditional calibration method is generally to obtain multiple images of the calibration object in different poses, and then use them as the input of the calibration algorithm for calibration. Therefore, the calibration pose has a greater impact on the calibration effect. [10,13] all pointed out some poses that can lead to bad results.

One of the main goals of interactive calibration is to help beginners avoid them; in addition, the difference in interactive calibration is that the system rather than humans determines the pose of the calibration object, that is, what orientation and where the calibration object should be placed, which frees beginners from the trouble of choosing a pose. Stable calibration can also be completed by learning specific calibration theory.

The existing interactive methods[14,15] are based on subjective experience to set a candidate pose set, and then formulate a pose scoring strategy, so that when needed, the best pose with the highest score can be selected. Moreover, the user-guided solution requires users to spend a lot of time to adjust the orientation of the calibration board; this paper designs a pose search algorithm framework based on the entire pose space, which can flexibly match different pose scoring and initialization strategies, and select the optimal pose in the entire pose space. The result is more reliable and accurate. This makes the development of new interactive methods simpler and easier..

**3. Calibration process**

This paper proposes an interactive camera calibration method based on search algorithm and pose decomposition, and designs a pose search algorithm to solve the optimal pose in the entire pose space. The calibration quality under different initial solutions and different loss function quantitative evaluation factors is analyzed, and the initial solution and loss function required for the optimal calibration effect are given. A step-by-step guidance method is designed. By decomposing the desired pose and displaying it, supplemented by text and rotation direction diagram for user guidance, it greatly reduces the burden on users and makes the pose realization faster and easier.

The process of the calibration method in this article is shown in Figure 3:

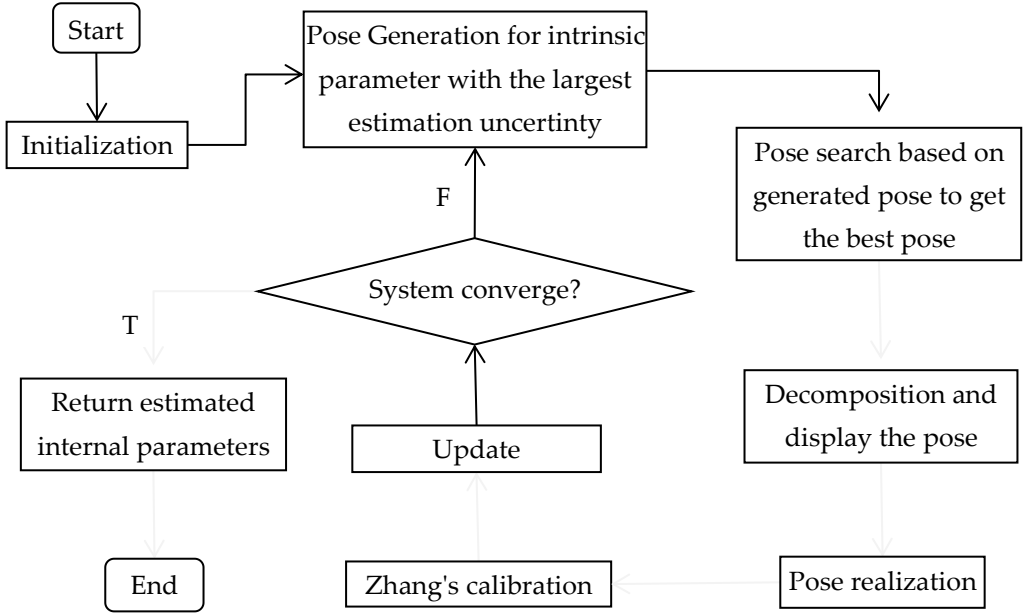

**Figure 3.** Calibration flow chart.

*3.1. Startup strategy*

At the beginning of calibration, the system does not contain any prior information of the camera internal parameters. In order to ensure the effect of the initial pose, we adopted the same startup strategy as[14,15]. Set initial pose $p = [45, 0, 0, 0, 0, z]$, where z is a preset value. Place the calibration board in front of the camera and gradually move it backwards until the entire calibration pattern is completely visible in the image. Estimate the distance z from the calibration board to the camera at this time. In the start-up phase, the system uses a limited camera model with no distortion and fixed optical center projection position at the center of the image, so that the scale factor $(\alpha, \beta)$ on the horizontal and vertical axes of the image can be estimated by calibrating only a single image for quick initialize internal parameters. In order to make the system initialization process more reliable, the real-time image is calibrated, and the estimated internal parameters are replaced every time a new internal parameter with lower reprojection error is found, until the user confirms to capture a calibration image.

After the system is started, the complete camera model is restored, and acquisition and realization of the remaining optimal poses are started until the system converges.

*3.2. Pose generation*

It is generally believed that a successful calibration must constrain each internal parameter, so Pose Selection[14] proposed a targeted pose generation strategy, which aims to constrain the internal parameter with the largest estimation uncertainty to generate the optimal pose. It groups the internal parameters. Apply different pose generation strategies, grouped as follows:

$$C_K = [\alpha, \beta, u_0, v_0]; C_\Delta = [k_1, k_2, k_3, p_1, p_2]$$

The goal of the group is to maximize the spread angle between the image plane and the calibration object. The pose part is a fixed value selected in advance. It should make the entire calibration object completely visible in the field of view. To avoid the calibration plane being parallel to the image, the pose generation formula is as follows:

$$p = \begin{cases} [0, \theta_{i1}, \pi/8, 0, 0, z]; & c \in [\alpha, u_0] \\ [\theta_{i2}, 0, \pi/8, 0, 0, z]; & c \in [\beta, v_0] \end{cases} \quad (10)$$

$$\begin{cases} \theta_1 = -70°; \\ \theta_2 = 70°; \\ \theta_i = (\theta_{i-1} + \theta_{i-2})/2; \ i > 2, i \in Z \end{cases} \quad (11)$$

where i represents the number of generations, and c represents the internal parameters that need to be constrained.

The goal of the group is to increase the sampling of the most distorted area, so its pose is generated as follows:

Generate a distortion map based on the current calibration result (the value at each position represents the deviation caused by the distortion coefficient acting on that point)

Find the rectangular area with the largest distortion in the picture in the form of sliding window

Perform pose estimation on the area to get the pose

*3.3. Pose search*

3.3.1. Algorithm definition

The search algorithm[24,25] is often used to find the optimal solution in a larger solution space and requires the following elements:

the definition of the solution and the solution space; (2) the error function to measure the quality of the solution; (3) the initial value of the solution; (4) the calculation strategy of the adjacent solution; (5) the strategy for updating the optimal solution; (6) the termination condition.

The pose search algorithm in this paper is based on the idea of simulated annealing[26,27], and its related definitions are as follows:

**Solution:** One solution is a pose that can be symbolized as $p = [xr, yr, zr, xt, yt, zt]^T$, which represents the transformation from the coordinate system of the calibration board to the camera coordinate system, where $xr, yr, zr$ represents the rotation angle under each coordinate axis, and $xt, yt, zt$ represents the translation in the direction of each coordinate axis.

**Solution space:** Because the rotation angle is too large, it is difficult to extract feature points, so the following constraints are made: $xr, yr, zr \in [-70°, 70°]$

Initial solution: Take the pose generated by the strategy in 3.2 as the initial solution;

**Adjacent solution:** Randomly select an element in the middle, sample it by Gaussian sampling[10] with mean and variance, copy the middle element, and get the adjacent solution after replacement;

**Solution update strategy:** Calculate the loss values of the current solution and adjacent solutions separately, and update the solution according to the following probability:

$$P(\Delta E) \begin{cases} 1 & \epsilon' < \epsilon \\ e^{-\Delta E/T} & \epsilon' >= \epsilon \end{cases} \qquad (12)$$

**Loss function:** Perform hypothetical calibration based on the system state and solution, obtain the hypothetical estimated value and variance of the internal parameters, and calculate the sum IOD[14] of all internal parameters as the loss value of the current solution;

$$IOD = \frac{\sigma^2}{C} \qquad (13)$$

$$\epsilon = SumIOD = \sum_i \frac{\sigma_i^2}{C_i} \qquad (14)$$

3.3.2. Search process

The process of pose search is shown in pseudo code 1. First, initialize the relevant parameters of the simulated annealing algorithm, such as the current solution, current temperature, termination temperature, cooling coefficient, the number of searches at each temperature, and then start a round of search. Repeat the following operations in each round of search: randomly select an adjacent solution of the current solution, calculate the loss function values of the two respectively, and update the solution based on the two loss values and the current temperature according to the update strategy of the solution. After the second operation, the temperature is lowered. When the current temperature is lower than the termination temperature, the search ends and the current solution is returned; otherwise, the next round of search process starts.

**Algorithm 1** Simulated Annealing
---
1: **function** SA( )
2:     Initialize $p, T_{now}, T_{min}, r, k$
3:     **while** $T_{now} > T_{min}$ **do**
4:         $i = 0$
5:         **while** $i < k$ **do**
6:             $p' = \text{GetNeighbor}(p)$
7:             $\Delta = \text{Cost}(p') - \text{Cost}(p)$
8:             $p = UpdateSolution(p, p', \Delta, T_{now})$
9:             $i = i + 1$
10:        **end while**
11:        $T_{now} = CoolDown(T_{now}, r)$
12:    **end while**
13:    **return** $p$
14: **end function**

### 3.4. Pose decomposition

After searching for the next optimal pose, the interactive system will map the calibration pattern to the camera image based on the current estimated internal parameters and optimal pose to inform the user of the ideal next placement position of the calibration board. Some auxiliary methods are usually used to help users place the calibration board more quickly and efficiently.

In the current guidance implementation scheme, AprilCal[15] highlights the overlapping part of the calibration pattern under the current pose and the optimal pose; Pose Selection[14] respectively visualized the three-dimensional coordinate system of the calibration board in two poses for the user as a reference to reduce the user's burden. But the visualized coordinate system needs to detect the calibration pattern of the imaging, it brings the problem of time delay.

On some older machines, the situation is more serious, which affects the user experience. In response to this problem, this paper proposes a step-by-step user guidance strategy based on pose decomposition. In the calibration process, lazy loading[16] is used instead of real-time detection of calibration patterns.

The target pose contains rich and complex three-dimensional spatial information. To achieve the target pose, good spatial thinking is required, which brings unnecessary burdens to users. As show in formula (15), the paper decomposes it into four poses, which are the results after first translation, then X-axis rotation, Y-axis rotation, and Z-axis rotation. The paper shows the position of the calibration objects after these transformations and the degree of transformation, which allows users to follow specific instructions and only need to care about the transformation in one dimension each time. The reduction of focus can greatly reduce the user's work the amount. In addition, only a set of transformation results need to be generated for each target pose, which can be saved as a reference for users. The figure shows the target pose before decomposition, and the figure shows the target pose displayed step by step. Finally, the figure shows the positive direction of rotation of each coordinate axis in the system coordinate system. The combination of the two makes the user guidance process simple, clear and efficient.

$$p = [xr, yr, zr, xt, yt, zt] \rightarrow \begin{cases} p1 = [0, 0, 0, xt, yt, zt] \\ p2 = [xr, 0, 0, xt, yt, zt] \\ p3 = [xr, yr, 0, xt, yt, zt] \\ p4 = [xr, yr, zr, xt, yt, zt] \end{cases} \quad (15)$$

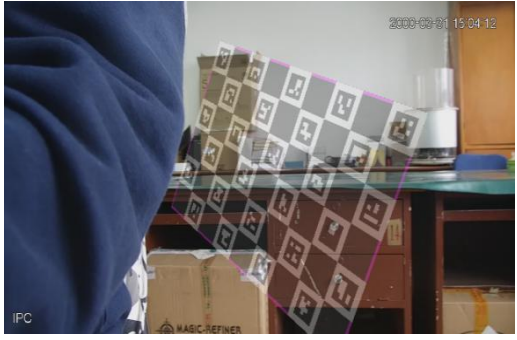
(a)

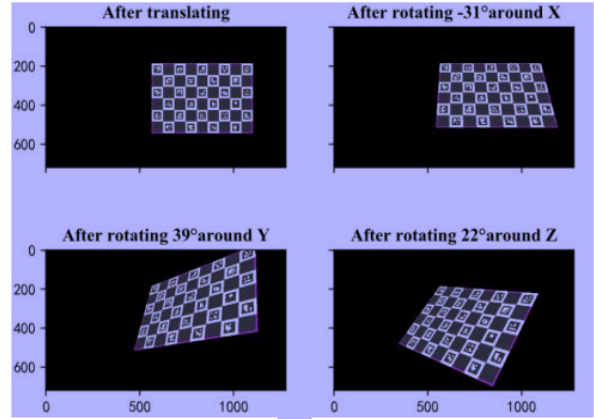
(b)

**Figure 4.** An example of pose decomposition. (a)The pose before decomposition can only rely on human experience to determine how to achieve it, (b)but after decomposition, this pose becomes quite intuitive and easy to understand: first translate to the center to the right, and then rotate 30 degrees around the negative half axis of the X axis, 39 degrees around the positive half axis of the Y axis, and 22 degrees around the positive half axis of the Z axis.

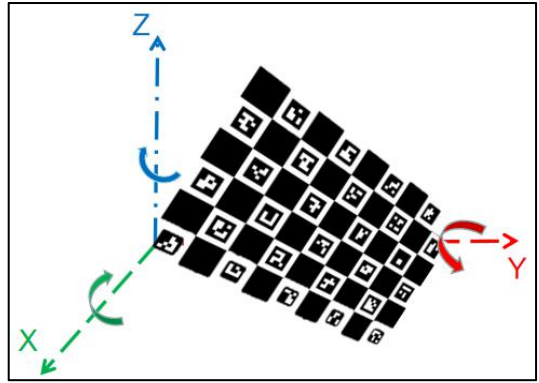

**Figure 5.** System coordinate system and the positive direction of rotation along each coordinate axis.

*3.5. System convergence*

Same as Pose Selection[14]: when the variance of the estimated internal parameter obtained from the last two calibrations does not change much, it is considered that the internal parameter has reached the convergence condition. The convergence determination formula is as follows:

$$r = \sigma_{n+1}^2/\sigma_n^2 \tag{16}$$

$$\phi = \begin{cases} 1-r <= \epsilon, & 1 \\ 1-r > \epsilon, & 0 \end{cases} \tag{17}$$

where n represents the number of calibration rounds; r represents the ratio of the variance of the internal parameters before and after the two rounds of calibration; $\epsilon$ represents the convergence threshold, which is generally set as 0.1. when $\phi$ is 1, the internal parameters converge.

When all internal parameters converge, the system converges, the algorithm ends, and the calibration result is returned.

**4. Evaluation**

This paper uses simulation data and measured data to evaluate the proposed method. The simulation data is to verify the feasibility of the pose search algorithm, and discuss the initial solution strategy and the influence of the loss function on the results; while the real data is compared with other interactive calibration methods to verify the effectiveness of the searched optimal pose and the improved user guidance strategy.

Before the experiment, we examined some calibration patterns, such as X-tag[28], Caltag[29], etc., and finally chose to use the self-identifying[30] pattern ChArUco[15,31]. In addition to the advantages of stable detection and anti-rotation interference, it is also because it is the pattern selected in most interactive calibration methods.. The pattern size is set to 9×6, each calibration picture can provide 40 feature points, and the grid width is 28mm.

*4.1. Simulation data*

In order to evaluate the effectiveness of the proposed method, 4 sets of comparative experiments were designed to verify the search algorithm, IOD[14] loss, the initial solution of the generated pose, and the improvement of the calibration effect brought by the combination of the three.

4.1.1. Evaluation indicators

This article uses the following two error indicators to measure the quality of calibration in a simulation scenario:
**SumIOD:** the sum of IOD[14] values of all internal parameters
**AbsRmsErr:** the absolute reprojection error. The real internal parameters are known in the simulation scene, so the coordinate deviation of the same 3D coordinate points after the remapping of the real internal parameters can be calculated

4.1.2. Experimental configuration

The properties of the simulated camera are set as follows:
$$[\alpha = 1068, \beta = 1073, u_0 = 635, v_0 = 355]$$
$$[k_1 = -0.0031, k_2 = -0.2059, k_3 = -0.0028, p_1 = -0.0038, p_2 = 0.2478]$$

The image resolution is that the calibration pattern contains a feature point, and Gaussian noise with a mean value of 0 and a variance of 0.1 is introduced to simulate the error of feature point detection in a real scene[10]. The number of standard positioning poses is 20. The search algorithm is configured as follows: The temperature is 1, the termination temperature is 0.1, the cooling coefficient is 0.7, and the number of searches at each temperature is 10 times. Therefore, at each initial pose, a maximum of 70 optimal pose updates may be performed. Each group of experiments will be repeated 20 times to take the average of the error.

In the following four sets of experiments, the abscissa represents the number of frames, and the ordinate represents one of the error indicators. Distinguish according to the mark at the top of the picture. Since the numerical changes before and after the error curve are more obvious, the comparison effect is not obvious, so we use the log function[32] with e as the base for processing.
**Investigate the impact of search algorithms**

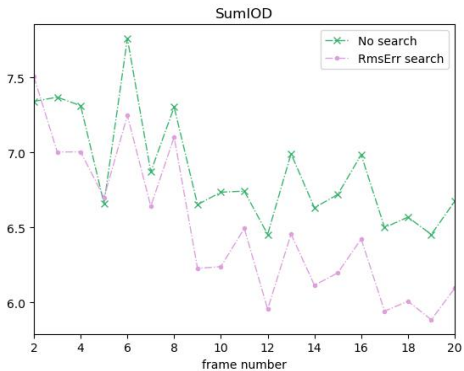

(a)

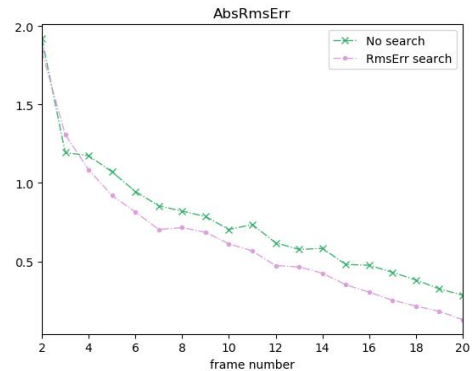

(b)

**Figure 6.** The red part in the figure below represents the curve of the calibration performance with the number of poses when the random effective pose is used as the target pose, and the blue part represents the calibration performance curve of the pose search result with the random effective pose as the initial solution and the classic RmsErr error [4,7] as the loss function. It can be seen that the calibration result after the search is reduced in uncertainty and reprojection error.

**RmsErr vs. MaxIOD, examine the impact of loss function**

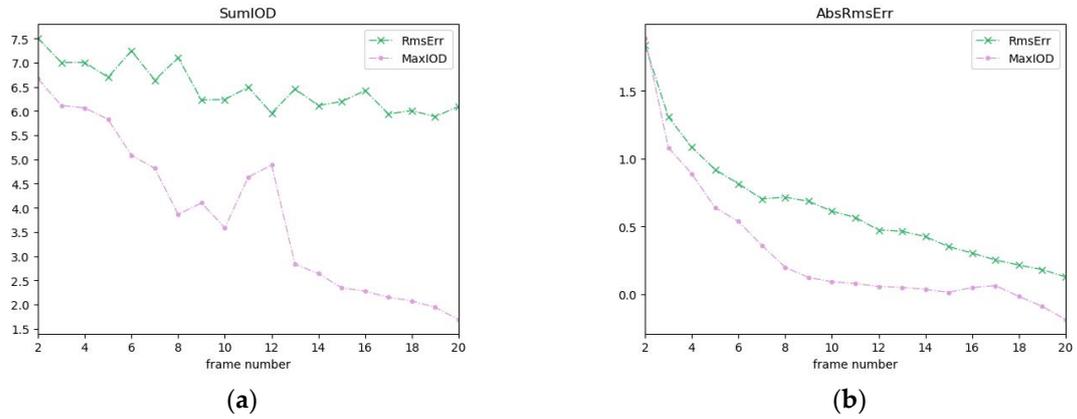

(a)           (b)

**Figure 7.** To verify the rationality of MaxIOD as a measure of calibration performance, the red and blue part represents the calibration performance curve when searching with RmsErr[4,7] and IOD[14] as the loss function. The search is all random effective pose as the initial solution. The results show that IOD[14] as a loss function has less volatility and is more reasonable.

**Random pose vs. Generated pose, examine the impact of initial solution**

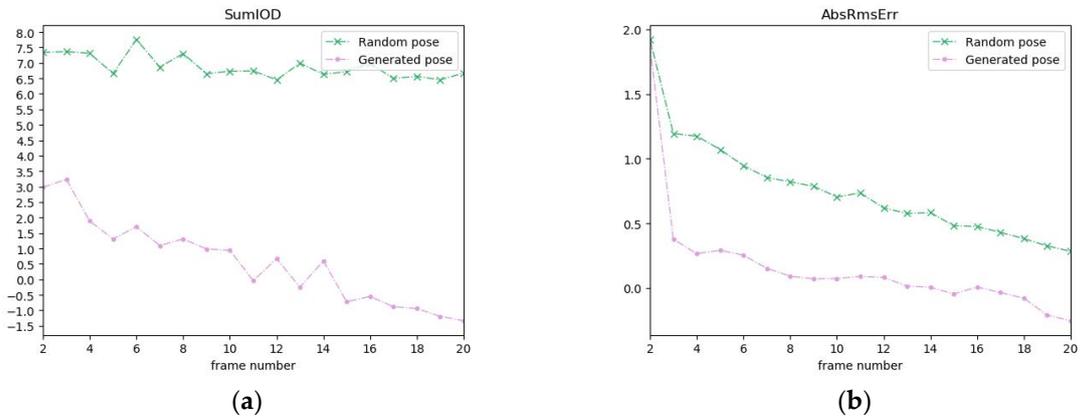

(a)           (b)

**Figure 8.** In order to examine the impact of different initial solution strategies on the search effect. Taking random pose and generated pose as the initial solutions, and the same loss function RmsErr[4,7]. The results show that when the generated pose in Pose Selection[14] is used as the initial solution, the evaluation of the calibration performance obtained by the search is far better than the random pose, and the volatility is smaller and more stable.

**Pose selection[14] vs. Pose search, test the effect of improvement**

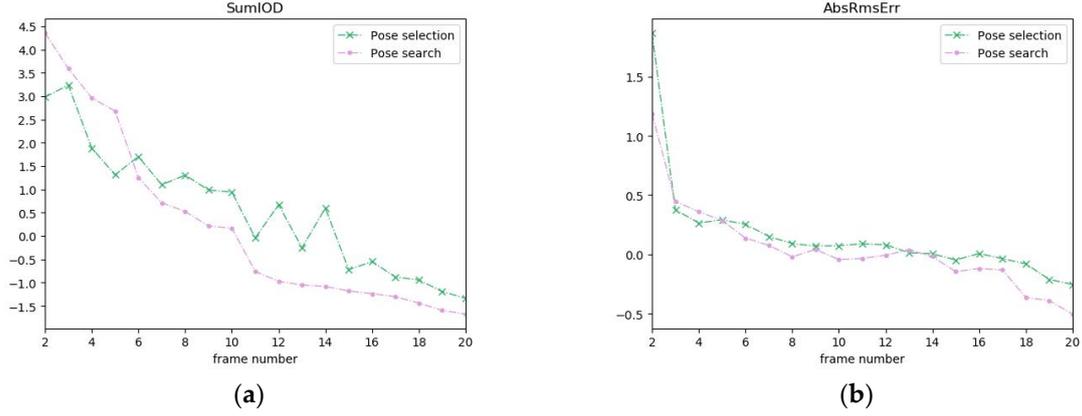

(a)          (b)

**Figure 9.** Comparing the scheme in this paper with Pose Selection[14], the results are shown in the figure. It can be seen that as the number of target positioning poses increases, after the number exceeds 5, the error of the system in this paper is lower than Pose Selection[14] under the three evaluation indicators.

*4.2. Real data*

In order to evaluate the method in this article on the measured data, 100 calibrated pictures with different positions and angles were captured in advance as the test set. All pictures were captured by Dahua network camera with a resolution of 1280×720px. The proposed method and Pose Selection[14] and OpenCV[33,34] without any pose constraints. The results show that, compared with the method in Pose Selection[14], the method in this paper has a 5.7% lower error when the number of calibration pictures is smaller. When the number of calibration pictures is only 86.7% of OpenCV[33,34], the error is reduced by 35.4%.

**Table 1.** The effect of each method is compared, each method is calibrated 5 times and the average error is taken.

| Method | Rms test | frames used | Rms train |
|---|---|---|---|
| Pose selection[14] | 0.4331(pixel) | 8.8 | 0.5139 |
| OpenCV[33,34] | 0.62253 | 9 | 0.43771 |
| Pose search | 0.4086 | 7.8 | 0.4704 |

In order to evaluate the effect of the proposed user guidance strategy, it was compared with the coordinate system reference strategy in Pose Selection[14], and 5 volunteers were invited to test. Volunteers have never performed camera calibration. They separately counted the time they took to achieve the same 10 poses under each method, similar to[35]: start timing after the tester completes the first pose. The experimental results are show in the table 3. It can be seen that most volunteers use the interactive guidance method mentioned in this article to place the calibration board in the expected pose faster. The second volunteer used the method in this article, and the time it took was very close to the method in Pose Selection[14]. The average time required by the interactive guidance method in this paper is 24.2% lower than that of the method in Pose Selection[14].

Then count the processing time of 5000 pictures under the two strategies. The processing content includes the detection of the calibration board and the projection of the target pose. The processing time of the real-time detection in Pose Selection[14] and the lazy loading mentioned in this article are 38.46ms and respectively. 25.24ms, the frame rate of the camera in this article is 25 frames/s, that is, the delay per second is 0.961s and 0.631s respectively. The processing time under lazy loading is reduced by 34.4%, which can greatly reduce the delay problem in the video. Here comes a better user experience.

**Table 2.** Comparison table of pose realization delay.

| Method | 5000 frames(s) | single frame(ms) | delay per second(s) |
|---|---|---|---|
| Pose selection[14] | 19.23 | 38.46 | 0.961 |
| Pose decomposition | 12.62 | 25.24 | 0.631 |

It can also be seen from the volunteers' feedback that the time delay greatly affects the user experience, which in turn affects the realization process of the pose guidance. In addition, the decomposed pose becomes clear and intuitive, with the coordinate axes provided in advance. By rotating the map in the positive direction, the entire guidance process becomes fast and efficient.

**Table 3.** Pose realization time comparison table.

| Method | volunteer1 | volunteer2 | volunteer3 | volunteer4 | volunteer5 |
|---|---|---|---|---|---|
| Pose selection[14] | 8:17 | 7:24 | 6:52 | 6:59 | 5:41 |
| Pose Decomposition | 7:19 | 7:28 | 3:42 | 4:07 | 5:08 |

## 5. Conclusions

In this paper, the research on the problem of the interactive internal parameter calibration of the camera is carried out, and an interactive calibration system based on pose search and pose decomposition--"Calibration Venus" is proposed. A pose search method is designed to find the optimal pose required for calibration. The pose search algorithm is designed to find the optimal pose that minimizes the uncertainty of the calibration estimation in the entire pose space. In order to improve the calibration efficiency, a step-by-step user guidance strategy based on pose decomposition is proposed. Our system is evaluated through simulation and measured data. The results show that our method has improved calibration effect and efficiency compared with the existing interactive method. It is foreseeable that more advanced search algorithms such as ant colony[36], evolutionary algorithm[37], etc. can obtain pose sets with better calibration effects.

The RmsErr[4,7] evaluation index used in the paper is a classic evaluation index, such as MaxRME[15], which helps to improve the stability of camera calibration. In addition, the current application field of interactive camera calibration is limited to a single camera. The internal parameter calibration of binocular[11] and multi-camera[38] needs further study. The next step will be from the above aspects.

## 6. Patents

**Author Contributions:** Conceptualization, Wentai Lei and Feifei Hou; methodology, Mengdi Xu; software, Mengdi Xu; validation, Wentai Lei, Mengdi Xu and Feifei Hou; formal analysis, Mengdi Xu; investigation, Wentai Lei; writing—review and editing, Feifei Hou; visualization, Wensi Jiang. All authors have read and agreed to the published version of the manuscript.

**Funding:** This research was funded by National Key R&D Program Safety Guarantee Technology of Urban Rail System, grant number 2016YFB1200402.

**Conflicts of Interest:** The authors declare no conflict of interest. The funders had no role in the design of the study; in the collection, analyses, or interpretation of data; in the writing of the manuscript, or in the decision to publish the results.**Author Contributions:** Conceptualization, Wentai Lei and Feifei Hou; methodology, Mengdi Xu; software, Mengdi Xu; validation, Wentai Lei, Mengdi Xu and Feifei Hou; formal analysis, Mengdi Xu; investigation, Wentai Lei; writing—review and editing, Feifei Hou; visualization, Wensi Jiang. All authors have read and agreed to the published version of the manuscript.

**Funding:** This research was funded by National Key R&D Program Safety Guarantee Technology of Urban Rail System, grant number 2016YFB1200402.

**Conflicts of Interest:** The authors declare no conflict of interest. The funders had no role in the design of the study; in the collection, analyses, or interpretation of data; in the writing of the manuscript, or in the decision to publish the results.